# GeoFitBench-LLM: A Benchmark for Assessing Large Language Models' Capabilities in Geospatial Data Requirements Elicitation for Geo-Analytic Tasks

Amirhossein Vahdat*, Thierry Badard* and Luc Lamontagne**

* Département des Sciences Géomatiques, Université Laval, QC, Canada
** Département d'informatique et de génie logiciel, Université Laval, QC, Canada

## Abstract

Large language models are becoming central to GeoAI systems, yet their evaluation rarely considers the elicitation of geospatial data requirements. This omission is critical as retrieval-oriented geo-analytic tasks cannot be supported reliably without first identifying the necessary datasets and constraints. GeoFitBench-LLM is introduced as a benchmark for assessing whether LLMs can infer the datasets, the supporting resources and the constraints required by geo-analytic requests in natural language. The benchmark is structured around two reasoning dimensions, Spatio-Temporal Relational and Semantic Grounding, reflecting long-standing challenges in geospatial information retrieval and related natural language tasks. Ground truth requirements are encoded in a fixed, provider-independent schema for reproducible evaluation of resource identification and attribute-level specification. Experiments on seven models under minimal prompting indicate that identifying the necessary resources is often achievable, with GPT-5.4 Thinking achieving the best result by recovering the complete set of resources in 66.67% of benchmark requests. Results further indicate that accurate fitness-for-use specification remains more difficult as inferential demands increase.

## 1. Introduction

Large language models (LLMs) are increasingly being incorporated into GeoAI systems that interpret natural language geospatial requests and support downstream analytical workflows (Ning et al. 2024, Ameen & Soilán 2026, Pierdicca et al. 2025). Recent evaluation efforts have assessed LLM performance on geospatial reasoning, multistep task completion, workflow synthesis, and code generation (Xu et al. 2025, Krechetova & Kochedykov 2025, Zhang et al. 2025). However, these benchmarks largely leave implicit a foundational capability for geospatial data retrieval: inferring the data requirements and constraints that must be identified to retrieve resources suited

Published in "Proceedings of the 1st International Conference on Geospatial Artificial Intelligence (GeoAI 2026) – Oral Presentation Papers", edited by Haosheng Huang and Nico Van de Weghe, GeoAI 2026, 3-6 June 2026, Ghent, Belgium.

This contribution underwent single-blind peer review based on the extended abstract.

for a given task. This omission matters because even when a model can propose a plausible workflow, the workflow cannot be executed meaningfully if the necessary datasets are missing, misidentified, or insufficiently specified (Ning et al. 2024).

The problem of eliciting geospatial data requirements predates LLMs and arises from longstanding challenges in geospatial information retrieval, geoportal discovery, and fitness-for-use assessment (Vockner & Mittlböck 2014, Hu et al. 2015, Li et al. 2017, Ziaimatin et al. 2020). This difficulty is especially visible in interdisciplinary contexts, where geospatial data are often used by domain specialists who are not necessarily GIS experts and who therefore tend to formulate requests in terms of the analytical or decision problem to be addressed rather than the datasets required to support it (Li et al. 2017, Ziaimatin et al. 2020). More broadly, geo-analytic requests are commonly framed around an intended analytical outcome rather than an explicit specification of the geospatial datasets required to achieve it. The central challenge, therefore, lies in recovering the latent data needs implied by the request rather than merely matching keywords to dataset descriptions. Effective support for such requests depends on accurate semantic grounding and spatio-temporal interpretation of analytical intent (Vockner & Mittlböck 2014, Hu et al. 2015, Li et al. 2017, Ziaimatin et al. 2020).

This work focuses specifically on geospatial data requirements elicitation. The central question is whether a large language model can infer which input and intermediate geospatial resources are needed for a natural language geo-analytic task and specify them in a form that is operationally useful for retrieval. GeoFitBench-LLM is introduced as a benchmark for assessing whether LLMs can interpret the analytical intent of geo-analytic requests and translate that intent into an explicit, provider-independent specification of the required data. By decoupling the request-understanding stage from concerns such as workflow execution, portal-specific retrieval, API usage, and tool orchestration, the benchmark focuses specifically on geospatial data requirements elicitation. GeoFitBench-LLM therefore supports not only model comparison but also a structured assessment of how well LLMs address long-standing challenges in geospatial information retrieval when the required data must be inferred rather than explicitly stated.

# 2. Towards Fitness-for-Use-Aware Geospatial Data Retrieval from Natural Language Requests

Natural language geospatial requests rarely express data needs in a complete retrieval-ready form. They are more often formulated as analytical objectives, while the necessary datasets, supporting resources, and operational constraints remain implicit. The central challenge, therefore, is not only to retrieve relevant data, but to infer the data requirements that render the requested analysis operationally feasible (Li et al. 2017, Hu et al. 2015). In this work, these recurring difficulties are organized into two complementary reasoning dimensions, Spatio-Temporal Relational and Semantic Grounding, which correspond to major challenges long recognized in geospatial information retrieval, geoportal discovery, geocoding, and natural language access to spatial data systems (Vockner & Mittlböck 2014, Hu et al. 2015, Pouliquen et al. 2006). The Spatio-Temporal Relational dimension captures cases in which data requirements are shaped by spatial or temporal structure in the request, whereas Semantic Grounding captures cases in which correct retrieval depends on lexical, conceptual, or referential interpretation. *Table 1* summarizes the criteria used in GeoFitBench-LLM.

**Table 1.** Benchmark reasoning dimensions and evaluation criteria

| Dimension | Identifier | Criterion | Retrieval challenge |
|---|---|---|---|
| Spatio-Temporal Relational | R1 | Spatial relation reasoning | Resources are defined through spatial relations such as containment, proximity, intersection, or buffering rather than by explicit name. |
| | R2 | Temporal relation reasoning | Resource suitability depends on correct temporal alignment with specified dates, periods, or events. |
| | R3 | Comparative reasoning | Requests require parallel and comparable evidence across places, spatial units, or time intervals under a shared analytical frame. |
| | R4 | Aggregation reasoning | The task depends on totals, counts, averages, or other summaries over spatial, temporal, or categorical units. |
| | R5 | Multi-hop relational reasoning | Data requirements emerge through chained dependencies or intermediate constraints. |
| | R6 | Granularity and hierarchical reasoning | Retrieval depends on reasoning across nested spatial or temporal levels. |
| | R7 | Network reasoning | Accessibility, flow, or distance are defined by network structure rather than geometric proximity. |
| Semantic Grounding | S1 | Synonymy and vocabulary mismatch | Relevant resources may remain undiscovered because request language differs from metadata terminology. |
| | S2 | Polysemy and homonymy | Context dependent terms may redirect retrieval toward the wrong type of resource. |
| | S3 | Acronyms and domain shorthand | Abbreviated technical expressions must be expanded correctly to recover intended requirements. |
| | S4 | Units, measurement scales, and thresholds | Quantities, units, and thresholds directly constrain resource suitability. |
| | S5 | Temporal reference and recency interpretation | Relative expressions such as current or recent must be resolved into operational time requirements. |
| | S6 | Feature property distinction | The request may name a variable or property rather than the broader dataset class that contains it. |
| | S7 | Implicit spatial verbs and constraints | Spatial meaning is expressed indirectly and must still be recovered for retrieval. |
| | S8 | Concept granularity and scope | Literal interpretation may yield resources that are too broad, too narrow, or otherwise misaligned. |
| | S9 | Place name and boundary disambiguation | Ambiguous geographic names or extents can propagate directly into incorrect data selection. |

# 3. GeoFitBench-LLM

The current version of GeoFitBench-LLM contains 48 benchmark items and is organized along two orthogonal design axes. The first axis is task difficulty and is structured into three levels:

- D1: direct dataset identification, in which the required resource is explicitly named or can be recovered with minimal inference from the request;
- D2: spatio analytical data inference, in which the request expresses an analytical objective and the model must infer the necessary data themes, supporting layers, or operationally required companion resources;
- D3: domain conceptual data inference, in which the request depends on modeled, derived, or domain mediated phenomena and the model must infer the conceptual inputs required to construct the requested evidence.

The second axis captures the type of reasoning challenge, namely the Spatio-Temporal Relational and Semantic Grounding dimensions introduced in Section 2. Each reasoning criterion is instantiated across the three difficulty levels, yielding 21 Spatio-Temporal Relational and 27 Semantic Grounding benchmark items. This design makes it possible to evaluate not only overall performance, but also how model behavior varies across distinct geospatial retrieval difficulties and increasing levels of inferential demand.

The ground truth is expressed as a normalized and provider-independent specification of the geospatial resources required to support the requested analysis, rather than as provider-specific products or catalog entries. To make these requirements explicit and machine-interpretable, GeoFitBench-LLM uses a fixed structured schema aligned with discovery-oriented concepts from ISO 19115-1 and ISO 19115-2 (ISO 2014, ISO 2019). For each required resource, the schema records resource name, role, theme, area of interest, time coverage, data type, other_attributes_include, spatial resolution or scale, temporal resolution, sensor or spectrum or bands, spatial reference, and data format. This representation enables evaluation not only of whether the correct resource types are identified, but also of whether their essential characteristics are specified in a manner consistent with the analytical task. The field other_attributes_include is used to specify additional mandatory dataset attributes or attribute level constraints necessary for the task to be valid or feasible but not covered by the core schema fields and not inherently guaranteed by selecting a correct dataset of the corresponding resource type. *Figure 1* illustrates a representative GeoFitBench-LLM item targeting R5 multi-hop relational reasoning, where successful requirement elicitation depends on inferring the datasets needed to identify hospitals in Montreal within 300 metres of main artery roads, determine the neighbourhoods containing those hospitals, and then identify the schools located within those neighbourhoods.

**$D_2$–$R_5$–$Q_1$:** "Obtain the datasets needed to find schools in the city of Montreal that are located within neighbourhoods containing hospitals that are themselves within 300 metres of main arteries roads."

**Ground-truth requirement representations**

{
"required_resources": [
{"resource_name": "School locations dataset", "role": "provide school locations used to identify schools in Montreal that are located within neighbourhoods containing hospitals that are themselves within 300 metres of roads classified as main artery roads", "theme": "education facilities / schools", "AOI": "City of Montreal", "time_coverage": "not specified", "data_type": "vector points", "other_attributes_include": [{"name": "not specified","value": "not specified"}], "spatial_resolution_or_scale": "not specified", "temporal_resolution": "not applicable", "sensor_spectrum_bands": "not applicable", "spatial_reference_CRS": "not specified", "data_format": "not specified"},
{"resource_name": "Neighbourhood boundary dataset", "role": "provide neighbourhood polygons used to identify neighbourhoods in Montreal that contain hospitals which are within 300 metres of roads classified as main arteries roads", "theme": "neighbourhoods / administrative units", "AOI": "City of Montreal", "time_coverage": "not specified", "data_type": "vector polygons", "other_attributes_include": [{"name": "not specified","value": "not specified"}], "spatial_resolution_or_scale": "not specified","temporal_resolution": "not applicable", "sensor_spectrum_bands": "not applicable", "spatial_reference_CRS": "not specified", "data_format": "not specified"},
{"resource_name": "Hospital locations dataset", "role": "provide hospital locations used to identify hospitals in Montreal that are within 300 metres of roads classified as main arteries roads and to determine which neighbourhoods contain those hospitals", "theme": "health facilities / hospitals", "AOI": "City of Montreal", "time_coverage": "not specified", "data_type": "vector points or vector polygons", "other_attributes_include": [{"name": "not specified","value": "not specified"}], "spatial_resolution_or_scale": "not specified", "temporal_resolution": "not applicable", "sensor_spectrum_bands": "not applicable", "spatial_reference_CRS": "not specified", "data_format": "not specified"},
{"resource_name": "Road network dataset", "role": "provide road geometries used to identify roads classified as main arteries roads and determine which hospitals in Montreal are within 300 metres of those roads", "theme": "transportation / road network", "AOI": "City of Montreal", "time_coverage": "not specified", "data_type": "vector lines", "other_attributes_include": [{"name": "road_classification_attribute","value": "must identify main arteries roads"}], "spatial_resolution_or_scale": "not specified", "temporal_resolution": "not applicable", "sensor_spectrum_bands": "not applicable", "spatial_reference_CRS": "not specified", "data_format": "not specified"}]
}

**Figure 1.** Example of a GeoFitBench-LLM benchmark item

Candidate outputs are generated under a minimal system prompt that instructs the model to act as a geospatial analyst and output only valid JSON in the benchmark schema. The prompt specifies one object per required resource, excludes derived outputs, prohibits provider-specific identifiers unless explicitly requested, and uses controlled placeholders for unconstrained or inapplicable fields. Evaluation is then performed using a separate LLM-as-a-judge prompt that compares each candidate answer against the benchmark ground truth using explicit parsing, resource alignment, and counting rules. In the judge protocol, alignment is based primarily on resource name and role, while theme, area of interest, time coverage and data type are used mainly as contradiction checks.

Model outputs are evaluated using two complementary metric families: Dataset Identification (DI) and Attribute Accuracy (AA). DI measures whether the required resources are correctly identified, whereas AA measures whether the required attributes of aligned resources are correctly specified. In addition to strict binary pass rates, the benchmark reports precision, recall, and F1 based partial credit measures, allowing it to distinguish between complete success, partial recovery of the required resource set, and errors in requirement specification. This separation is essential because a model may identify the correct resources while still failing to specify the attributes that determine their fitness for the intended analytical use. GeoFitBench-LLM therefore supports both comparative evaluation across models and diagnostic analysis of where failures arise in geospatial data requirements elicitation.

# 4. Experiments

GeoFitBench-LLM was evaluated under the minimal prompting conditions described in Section 3. In this setting, models received only the benchmark task definition and the required JSON schema, without worked examples or decomposition guidance. Seven models were evaluated: GPT-5.4 Thinking, GPT-5.4, Gemini 3.1 Pro, Gemini 2.5 Flash, GPT-OSS 20B, DeepSeek-R1 14B, and Gemma 3n. Performance is reported primarily using Dataset Identification for the full set of 48 benchmark items and for each reasoning subset, while Attribute Accuracy is reported to assess the quality of resource-level specification.

*Table 2* reports overall model performance in terms of DI_binary, DI_F1, and AA_F1 across the full benchmark, as well as DI_binary results for the Spatio-Temporal Relational and Semantic Grounding subsets. GPT-5.4 Thinking achieves the strongest overall performance, followed by GPT-5.4 and Gemini 3.1 Pro, while GPT-OSS 20B is the strongest open-source model among those evaluated.

**Table 2.** Overall benchmark performance across models under minimal prompting

| **Model** | **Combined DI_binary** (%) | **Spatio-temporal DI_binary** (%) | **Semantic DI_binary** (%) | **Combined DI_F1** (%) | **Combined AA_F1** (%) |
|---|---|---|---|---|---|
| GPT-5.4 Thinking | 66.67 | 52.38 | 77.78 | 80.81 | 78.23 |
| GPT-5.4 | 54.17 | 61.90 | 48.15 | 77.77 | 72.47 |
| Gemini 3.1 Pro | 50.00 | 57.14 | 44.44 | 79.49 | 77.36 |
| GPT-OSS 20B | 47.92 | 52.38 | 44.44 | 75.07 | 72.70 |
| Gemini 2.5 Flash | 39.58 | 42.86 | 37.04 | 73.52 | 73.48 |
| DeepSeek-R1 14B | 31.25 | 33.33 | 29.63 | 58.67 | 44.73 |
| Gemma 3n | 22.92 | 19.05 | 25.93 | 47.77 | 37.07 |

*Figure 2* illustrates average DI_binary by criterion for the Spatio-Temporal Relational dimension (*Figure 2a*) and the Semantic Grounding dimension (*Figure 2b*). Comparative reasoning and temporal relation reasoning are among the weakest areas in the Spatio-Temporal Relational dimension, whereas place name and boundary disambiguation, feature property distinction, and temporal reference and recency interpretation are among the weakest areas in Semantic Grounding. By contrast, aggregation and network reasoning, as well as synonymy and units and thresholds, are handled more reliably.

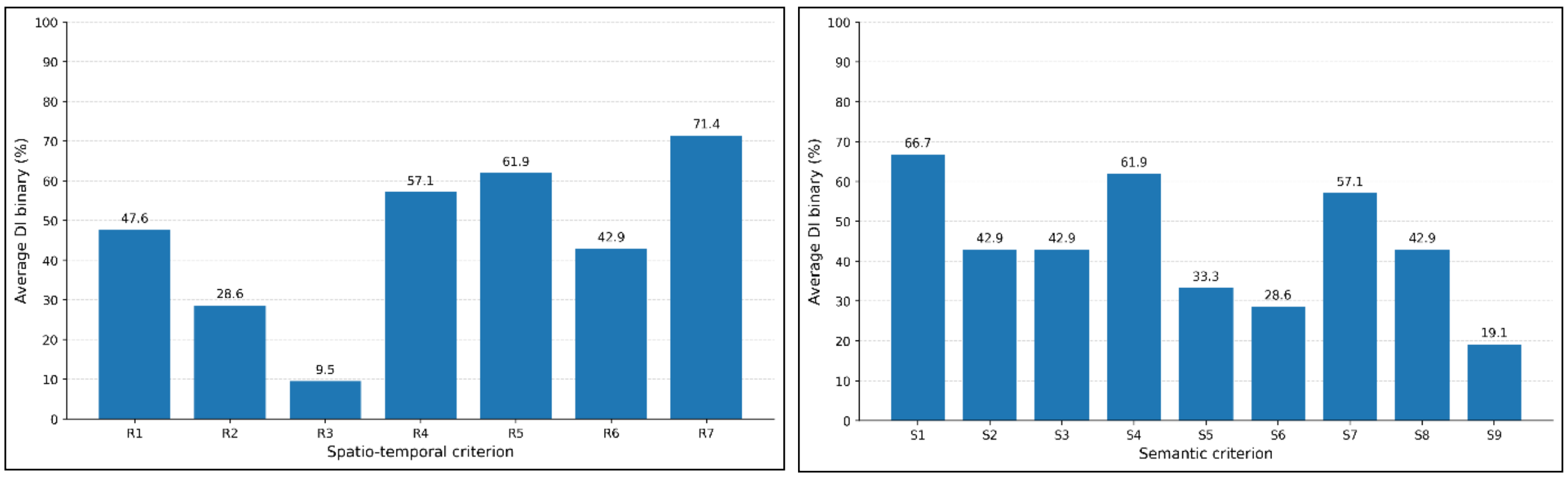

**Figure 2.** Average DI_binary by criterion across the two benchmark dimensions: (a) Spatio-Temporal Relational; (b) Semantic Grounding.

*Figure 3* reports results by difficulty level and shows a clear decline from D1 to D3, indicating that models are substantially more reliable for explicitly recoverable data needs than for inferred or domain conceptual requirements.

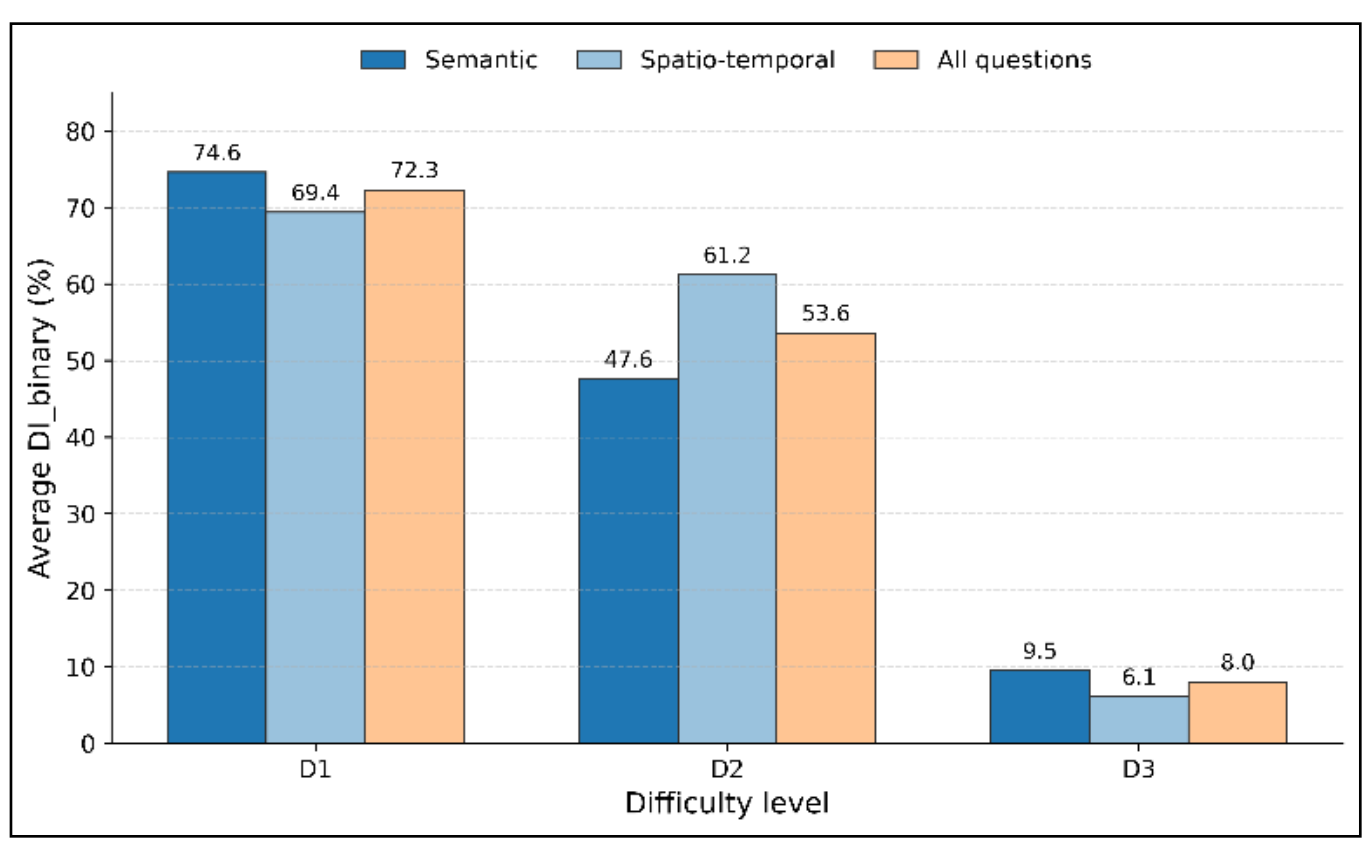

**Figure 3.** Average DI_binary by difficulty level for the Spatio-Temporal Relational subset, the Semantic Grounding subset, and the combined benchmark

# 5. Conclusions and Future Work

In this work, GeoFitBench-LLM was introduced as a benchmark for evaluating whether large language models can infer geospatial data requirements from natural language geo-analytic requests. The benchmark was designed around two complementary reasoning dimensions, Spatio-Temporal Relational and Semantic Grounding, and three difficulty levels ranging from direct dataset identification to more advanced domain-level conceptual inference. This design enables a structured evaluation of retrieval-oriented geospatial requirement inference across different forms of reasoning difficulty.

The experimental results show that geospatial data requirements elicitation remains a challenging capability for current LLMs. Performance varies substantially across models and declines as the inferential demands of the task increase, indicating that models are considerably more reliable when required resources are explicitly recoverable from the request than when they must be inferred from analytical intent. Overall, GeoFitBench-LLM provides a systematic basis for assessing a foundational yet under-evaluated capability in GeoAI and for identifying which models are better suited to retrieval-oriented geospatial tasks.

Future work will extend the benchmark in several directions. First, future work will expand the benchmark items to cover a broader range of geospatial tasks and application domains, while involving additional geospatial and domain specialists in item design and ground-truth validation, with the aim of improving coverage, strengthening validation, and enabling more fine-grained diagnosis of recurring elicitation failures. Another priority is to evaluate the same models under alternative prompting strategies, including few-shot prompting, Decomposition (DECOMP), and chain-of-thought prompting, in order to examine how far prompt design can improve geospatial data requirements elicitation. Another important direction is a more detailed error analysis of benchmark outputs, particularly to identify which schema attributes are most challenging for LLMs to specify correctly and which types of requirement inference remain most error-prone across models. Such analyses can help clarify whether current limitations arise primarily from resource identification, attribute specification, or prompt sensitivity, and can in turn guide the development of more effective prompting strategies and future model improvements.

## Acknowledgement

We acknowledge the support of the Natural Sciences and Engineering Research Council of Canada (NSERC) through the Discovery Grant.